\documentclass{article}

\usepackage{microtype}
\usepackage{graphicx}
\usepackage{subfigure}
\usepackage{booktabs} 

\usepackage{hyperref}
\usepackage[]{mdframed}

\usepackage{amsmath,amsfonts,bm}

\newcommand{\newterm}[1]{{\bf #1}}

\def\eqref#1{equation~\ref{#1}}

\def\1{\bm{1}}

\DeclareMathAlphabet{\mathsfit}{\encodingdefault}{\sfdefault}{m}{sl}
\SetMathAlphabet{\mathsfit}{bold}{\encodingdefault}{\sfdefault}{bx}{n}

\def\gG{{\mathcal{G}}}

\def\gM{{\mathcal{M}}}

\def\gS{{\mathcal{S}}}

\def\gX{{\mathcal{X}}}
\def\gY{{\mathcal{Y}}}

\def\sN{{\mathbb{N}}}

\def\sR{{\mathbb{R}}}

\newcommand{\E}{\mathbb{E}}

\newcommand{\R}{\mathbb{R}}

\newcommand{\pomg}{\mathcal{M}}
\newcommand{\states}{S}
\newcommand{\actions}[1][]{\ifthenelse{\equal{#1}{}}{A}{A_{#1}}}
\newcommand{\action}[1][]{\ifthenelse{\equal{#1}{}}{\bm a}{a_{#1}}}
\newcommand{\actiont}[1][]{\ifthenelse{\equal{#1}{}}{{\bm a}_t}{a_{#1,t}}}
\newcommand{\transfunc}{T}
\newcommand{\rewardfunc}{r}

\newcommand{\obs}{\Omega}
\newcommand{\obfunc}{O}
\newcommand{\discount}{\gamma}
\newcommand{\type}[1][]{\ifthenelse{\equal{#1}{}}{\theta}{\theta_{#1}}}
\newcommand{\typet}[1][]{\ifthenelse{\equal{#1}{}}{{\bm \theta}_t}{\theta_{#1,t}}}
\newcommand{\types}[1][]{\ifthenelse{\equal{#1}{}}{\Theta}{\Theta_{#1}}}
\newcommand{\br}{\mathcal{B}\mathcal{R}}

\usepackage{xifthen}
\usepackage{bbm}
\usepackage{color}
\usepackage{seqsplit}
\usepackage{xargs}
\usepackage{xstring}
\DeclareMathAlphabet\mathbfcal{OMS}{cmsy}{b}{n}
\usepackage{commands}

\allowdisplaybreaks

\usepackage[accepted]{icml2024}

\usepackage{amsmath}
\usepackage{amssymb}
\usepackage{mathtools}
\usepackage{amsmath, amssymb, amsthm}
\usepackage{wrapfig}

\usepackage{amsthm}

\usepackage[capitalize,noabbrev]{cleveref}

\theoremstyle{plain}
\newtheorem{theorem}{Theorem}[section]

\theoremstyle{definition}
\newtheorem{definition}[theorem]{Definition}

\theoremstyle{remark}

\usepackage[textsize=tiny]{todonotes}

\def\game{Social Environment Design Game}
\def\framework{Social Environment Design}

\usepackage{xcolor} 
\newcount\Comments  
\newcommand{\kibitz}[2]{\ifnum\Comments=1\textcolor{#1}{#2}\fi}

\icmltitlerunning{Position: Social Environment Design}

\begin{document}

\twocolumn[
    \icmltitle{Position: Social Environment Design Should be Further Developed for AI-based Policy-Making}



    \icmlsetsymbol{equal}{*}
    \begin{icmlauthorlist}
        \icmlauthor{Edwin Zhang}{harvard,founding}
        \icmlauthor{Sadie Zhao}{harvard}
        \icmlauthor{Tonghan Wang}{harvard}
        \icmlauthor{Safwan Hossain}{harvard}
        \icmlauthor{Henry Gasztowtt}{ox}
        \icmlauthor{Stephan Zheng}{asari}
        \icmlauthor{David C. Parkes}{harvard}
        \icmlauthor{Milind Tambe}{harvard,google}
        \icmlauthor{Yiling Chen}{harvard}
    \end{icmlauthorlist}

    \icmlaffiliation{harvard}{Harvard University}
    \icmlaffiliation{ox}{Oxford University}
    \icmlaffiliation{founding}{Founding}
    \icmlaffiliation{google}{Google Research}
    \icmlaffiliation{asari}{Asari AI}

    \icmlcorrespondingauthor{Edwin Zhang}{ezhang@g.harvard.edu}

    \icmlkeywords{Machine Learning, ICML}

    \vskip 0.3in

] 

\makeatletter\def\Hy@Warning#1{}\makeatother
\printAffiliationsAndNotice{}  
\newcommand\figlen{.48}

\begin{abstract}
    Artificial Intelligence (AI) holds promise as a technology that can be used to improve government and economic policy-making. This paper proposes a new research agenda towards this end by introducing \textbf{\framework}, a general framework for the use of AI in automated policy-making that connects with the Reinforcement Learning, EconCS, and Computational Social Choice communities. The framework seeks to capture  general economic environments, includes voting on policy objectives, and gives a direction for the systematic analysis of government and economic policy through AI simulation. We highlight key open problems for future research in AI-based policymaking. By solving these challenges, we hope to achieve various social welfare objectives, thereby promoting more ethical and responsible decision making.
\end{abstract}

\section{Introduction}
Economic policy formulation is a domain fraught with complexity, with traditional economic models providing limited foresight into the outcomes of policy decisions. Policy-makers must not only understand the immediate implications of individual policies but also their aggregate and long-term effects. In addition, human policy-maker incentives are oftentimes not aligned with the interests of the general public, and may instead prioritize special interests or reelection \citep{de_figueiredo_advancing_2014}. In light of this, AI-based approaches to policy design that can simulate economies and target different objectives, hold the potential for improved policy understanding and formulation \citep{zheng2020ai,koster2022humancentered}.~

\setlength{\intextsep}{0pt}%
\begin{figure}[t]
    \centering
    \includegraphics[width=\figlen\textwidth]{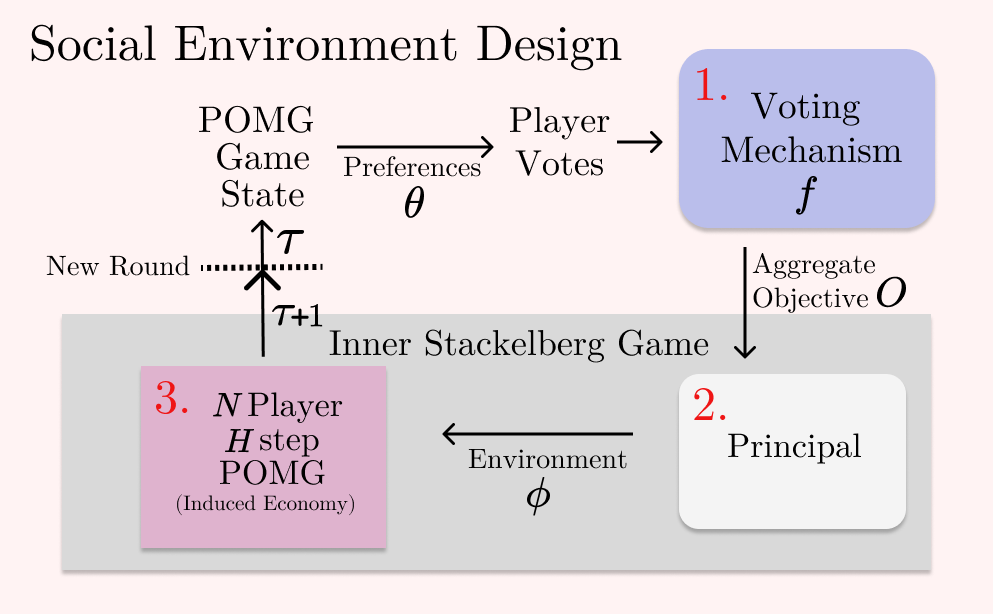}
    \vspace{-10pt}
    \caption{ {\em The proposed framework.} The process begins with voting, where human or AI players report preferences on social welfare objectives to a voting mechanism (1). This an objective for a Principal policy-maker, who designs a parameterized $N$-player Partially Observable Markov Game (POMG) (2). The players are the same as the voters.  This POMG unfolds over several timesteps $H$ (3). Following the POMG, game state information is extracted to initiate $n$ last POMG state used as the first game state of the new round. This whole process is repeated again in the next round.}
    \label{mainfig}
\end{figure}
Wra where AI is gaining increasing attention in  governments \citep{house_executive_2023,engstrom_government_2020}, it is timely to understand its potential influence on future policy-making. Ideally, such a  framework should satisfy the following desiderata:

\begin{enumerate}
    \item  \textbf{Alignment of policy-makers} to the values of  constituents, whilst ensuring fair and equitable representation \citep{barocas-hardt-narayanan}.
    \item  Sufficient \textbf{model expressivity} \citep{Patig2004MeasuringEI} to accurately represent the intricate governance structures found in the real world, capturing the subtleties and variances of socio-economic interactions.
    \item Balance expressiveness with \textbf{computational tractability}, making it feasible to scale to systems with a large number of agents.
    \item Build \textbf{theoretical understanding}, enabling systematic analysis and offering a new lens on complex economic models.
\end{enumerate}

In this paper, we propose a new  framework, Social Environment Design, that lays out an agenda in making progress towards these desiderata. In our framework, illustrated in \autoref{mainfig}, we suggest addressing the concern of a misaligned policy-maker with ``Voting on Values \citep{hanson2013shall}," coupled with a Principal policy-maker (or environment designer) who seeks to achieve  suggested policy goals. We capture the complexity of a general economic environment whilst maintaining computational tractability by modeling the economy as a Partially Observable Markov Game (POMG), which maintains a fixed observation space for each agent. Finally, we structure our framework as repeatedly finding Stackelberg Equilbria, enabling theoretical understanding
by allowing reduction to simpler subproblems.

We now state the position of this paper: \textbf{Social Environment Design should be further developed in order to enable AI-based policy-making.} Towards this end, we discuss several open problems of  practical and theoretical interest. By introducing this framework, we open a dialogue on AI’s application to economic and government policy design, aspiring to someday help leverage AI to assist policymakers in enhancing economic resilience and governance effectiveness.

In summary, we list our core contributions:

1) We propose the {\framework} framework to enable future research in AI-led policymaking in complex economic systems;

2) We release a  core implementation of our framework as a Sequential Social Dilemma Environment along with code; and

3) We provide a characterization of open problems, along with prospective solution concepts and algorithmic approaches to forward the dialogue on AI's application in economic policy design.

\section{Preliminaries}\label{sec:prelim}

Here we give some preliminaries on several foundational games and solution concepts. We start by introducing the foundational concept of a Stackelberg Game and Partially Observable Markov Game, and combine these concepts together into the Stackelberg Markov Game. We then introduce the Mechanism Design problem, which studies a related problem as Social Environment Design in a more constrained setting.

\begin{definition}
    A $(n+1)$-player \newterm{Stackelberg-Nash Game} $\gS=(n,\gX, \gY, \bm u)$ comprises one player called the \newterm{leader} and $n\in \sN\setminus\{0\}$ players called \newterm{followers}. In a Stackelberg-Nash game, the leader first commits to an action $\bm x\in \gX$ from action space $\gX\subset \sR^m$. Then, having observed the leader's action, each follower $i\in [n]$, responds with an action $y_i$ in their action space $\gY_i\subset \sR^m$. We define the followers' joint action space $\gY=\bigtimes_{i\in [n]} \gY_i$. We refer to a collection of actions $\bm y=(y_1, \cdots, y_n)\in \gY$ as a followers' action profile, and to a collection $(x,y)\in \gX\times \gY$ as an action profile.

    After all players choose an action, the leader receives payoff $u_o(\bm x, \bm y)\in \sR$, while each follower $i\in [n]$ receives payoff $u_i(\bm x, \bm y)\in \R$. Each player $i\in [n]$ aims to maximize her payoff, and the leader aims to maximize her payoff assuming the followers will best respond.


    Fixing the leader's action $\bm x \in \gX$, a Stackelberg Nash game $\gS$ induces a \newterm{lower-level Nash game} $\gG^{\gS}=(n,m, \gY, \bm u_{-0}(\bm x, \cdot))$ among the followers.
\end{definition}


\begin{definition}
    A \textbf{Partially Observable Markov Game} (POMG) $\pomg$ with $n$ agents is a tuple $(\states, \actions, \transfunc, \rewardfunc, \obs, \obfunc, \discount,\mu_0)$. Note that this game is also referred to as a Partially Observable Stochastic Game (POSG).
    Here,
    \begin{itemize}
        \item $\states$ is a shared state space for all agents;
        \item $\actions=\bigtimes_{i\in [n]} \actions_i$ is the joint action space;
        \item $\transfunc:\states \times \states \to  \Delta(\actions)$ is a stochastic transition function;
        \item $\rewardfunc: \states\times \actions \to \R^n$ is the reward function with $r=(r_1, \cdots, r_n)$;
        \item $\obs=\bigtimes_{i\in [n]} \obs_i$ is the joint observation space;
        \item $\obfunc: \states\times \actions \to \Delta(\obs) $ is the stochastic observation function;
        \item $\discount \in [0, 1)$ is a discount factor;
        \item $\mu_0\in \Delta(\states)$ is the initial state distribution.
    \end{itemize}

    An agent's behavior in this game is characterized by its policy $\pi_i: \obs \to \actions$, which maps observations to actions.
\end{definition}

We now combine this two foundational concepts together.

\begin{definition}
    A $(n+1)$-player \newterm{Stackelberg-Markov Game} $\gS=(n, \Phi, \Pi, \bm u)$ comprises one player called the \newterm{leader} and $n\in \sN\setminus\{0\}$ players called \newterm{followers}. In a Stackelberg-Markov game, the leader first commits to an action $\phi \in \Phi$ from action space $\Phi \subset \sR^{m}$ which induces a $n$-player \newterm{low-level (Partially Observable) Markov Game} $\gM^{\phi}=(\states, \actions^{\phi}, \transfunc^{\phi}, \rewardfunc^{\phi}, \obs^{\phi}, \obfunc^{\phi},\discount,\mu_0^{\phi})$. Then, having observed the leader's action, each follower $i\in [n]$, responds with an policy $\pi_i: \obs \to \actions_i$ in their policy space $\Pi_i$. We define the followers' joint policy space $\Pi=\bigtimes_{i\in [n]} \Pi_i$. We refer to a collection of policies $\pi=(\pi_1, \cdots, \pi_n)\in \Pi$ as a followers' policy profile.

    After all players choose an action, the leader receives payoff $u_o(\phi, \pi)\in \sR$, while each follower $i\in [n]$ receives payoff

    \begin{equation}\label{eqn:policy-objective}
        u_i(\phi, \pi)=\E^{\gM^{\phi}, \pi}[\sum_{t=0}^{\infty}\gamma^t r_i^{\phi}(s^t, a^t)] \in \R.
    \end{equation} 
    Each player $i\in [n]$ aims to maximize her payoff, and the leader aims to maximize her payoff assuming the followers will best respond.

    For all followers $i\in [n]$, we define the \newterm{$\delta$-best-response correspondence}
    \begin{equation}
        \begin{aligned}
            \br^{\delta}_i(\phi, \pi_{-i}) & =\{\pi_i\in \Pi_i \mid u_i(\phi, \pi) \\ &\geq
            \max_{\pi_i\in \Pi_i} u_i(\phi, (\pi_i, \pi_{-i}))-\delta\},
        \end{aligned}
    \end{equation}

    and the \newterm{joint $\delta$-best-response correspondence}
    $\br^{\delta}(\phi, \pi)=\bigtimes_{i\in [n]} \br^{\delta}_i (\phi, \pi_{-i})$.
\end{definition}

\begin{definition}
    A \newterm{$(\varepsilon, \delta)$-strong Stackelberg-Markov-Nash equilibrium (SSMNE)} in a Stackelberg-Markov game $\gS=(n,\Phi, \Pi, \bm u)$ is an action profile $(\phi^*, \pi^*)\in \Phi\times \Pi$ such that
    \begin{equation}
        \begin{aligned}
            u_0(\phi^*, \pi^*)\geq \max_{\phi\in \Phi} \max_{\br^{\delta}} u_0(\phi, \pi)-\varepsilon~\text{and} \\
            u_i(\phi^*, \pi^*)\geq \max_{\pi_i\in \Pi_i} u_i(\phi^*, (\pi_i, \pi_{-i}^*))-\delta,~\forall i\in [n].
        \end{aligned}
    \end{equation}
\end{definition}

We now introduce the Mechanism Design (MD) problem, which while somewhat similar in motivation to Social Environment Design (SED), is a very different setting. SED considers POMG environments rather than the single-shot mechanisms found in MD. Additionally, in MD the mechanism stays fixed, and the optimal mechanism should be found in closed-form. On the other hand, SED will iteratively attempt to find a better environment, and makes no claims around optimality or closed-formedness of the generated environment.

\begin{definition}
    A \newterm{(One-Shot) Mechanism Design} problem $\mechprob=(\nummagents, \mtypespace, \outcomespace, \mtype, \mutil, \objfunc, f)$ comprises of $\nummagents$ agents, each $\magent\in \magents$ owning a private type $\mtype[\magent] \in \mtypespace[\magent]$ from a set of possible types $\mtypespace[\magent]\subset \R^{\numactions}$.
\end{definition}

Our goal in Mechanism Design is to design some system known as mechanism which achieves some collective social outcome. In order to do so, we must first be able to capture the values and preferences of an agent. An agent's preferences over outcomes $\outcome\in \outcomespace$, for a set $\outcomespace$ of outcomes, can be expressed in terms of a utility function that is parameterized by the type. Let $\mutil[\magent](\outcome, \mtype[\magent])$ denote the utility of agent $\magent$ for outcome $\outcome\in \outcomespace$ given type $\mtype[\magent]$.
A strategy (which formally defines an agent's behavior) $\strategy[\magent]: \mtypespace[\magent] \to \mactionspace[\magent]$ chooses an action based on the given type.
Let $\maction[\magent]=\strategy[\magent](\mtype[\magent])\in \mactionspace[\magent]$ denote the action of agent $\magent$ given type $\mtype[\magent]$, where $\mactionspace[\magent]$ is the set of all possible actions available to agent $\magent$.

A \newterm{mechanism} $\mech=(\{\mactionspace[\magent]\}_{\magent\in \magents},\mechrule)$ defines the set of actions $\mactionspace[\magent]$ available to each agent $\magent$, and an outcome rule $\mechrule: \bigtimes_{\magent\in \magents} \mactionspace[\magent]\to \outcomespace$, such that $\mechrule(\maction)$ is the outcome implemented by the mechanism for action profile $\maction=(\maction[1],\cdots, \maction[\nummagents])$.
$\objfunc: \mechspace \times \bigtimes_{\magent\in\magents}\mactionspace[\magent] \to \R$ is a principal objective function, where $\objfunc(\mech, \maction)$ represents the expected utility/revenue of the principal when mechanism designer chooses mechanism $\mech$ and agents choose action profile $\maction$. Note that $\objfunc$ defines this collective social objective, also known as the social choice function.
The goal of the mechanism designer is to design a  mechanism $\mech=(\{\mactionspace[\magent]\}_{\magent\in \magents},g)\in \mechspace$ that maximizes $\objfunc(\mech, \maction^*)$, where strategy profile $\maction^*=(\maction[1]^*, \cdots, \maction[\nummagents]^*)$ is an (Nash \cite{nash1951non}, Bayesian-Nash \cite{john1968bayesian}, dominant-strategy \cite{laffont1982nash}) equilibrium to the game induced by $\mech$.


\section{Formal Definition of \game}\label{sec:framework}
\vspace{10pt}
\def\induced{Induced Economy}
\newcommand{\objective}{w}
\newcommand{\objectives}{W}


\begin{mdframed}
    
\begin{definition}\label{def:main-framework}
A \newterm{\game} $\mathcal{S}=(\Phi, P, \phi_0, D, \delta, \Theta, \mathcal{O}, f)$ is a one-leader-$n$-follower online\footnote{
   Here, online means that the Stackelberg-Markov Game is repeatedly played, with the first state of a new round made equal to the final state of the last round.} Stackelberg-Markov Game, where 
\end{definition}

\begin{itemize}
    \item  $ \Phi \subseteq \mathbb{R}^k$ is the principal action space;
    \item $P: \Phi \mapsto \pomg^{\phi}$ is a policy implementation map that maps from a principal action $\phi \in \Phi$ to a parameterized POMG $\pomg^{\phi}=(\states, \actions^{\phi}, \transfunc^{\phi}, \rewardfunc^{\phi}, \obs^{\phi}, \obfunc^{\phi}, \discount^{\phi}, \mu_0^{\phi})$;
    \item  $\phi_0 \in \Phi$ is some initial action;
    \item $D: \Phi \times \Phi \mapsto \mathbb{R}_{\geq0}$ is a divergence measure on the leader action space;
    \item $\delta > 0$ is the divergence constraint;
    \item $\Theta \subseteq \mathbb{R}^{(n + 1) \times m}$ is the type space; 
    \item $\objectives = \{\objective_{i}\}_{i\in [l]}$ is some set of predefined social welfare functions, where each $\objective$ maps $\Phi \times \Pi \mapsto \mathbb{R}$. We give examples of several possible choices of objectives below in \hyperref[sec:obj-examples]{Social Welfare Examples}. $\Pi$ here refers to the set of all possible policy profiles in the parameterized POMG;
    \item $f: \Theta \mapsto \objectives$ is a social choice function representing the voting mechanism.
\end{itemize}

\end{mdframed}

We define the {\game} formally as repeatedly finding a Stackelberg Equilibrium in a Markov Game \citep{gerstgrasser2023oracles,brero2022learning}, iterated over several rounds of voting.

At a high level, we frame the  economic design problem as a Stackelberg game between the policy designer and economic participants. The economic participants first vote for a given objective, or values to optimize for. Subsequently, the Principal (leader) attempts to maximize this objective by designing the rules of an economic system, which induce an environment for the participants. We model this environment as a Partially Observable Markov Game (POMG) with theparticipants as the agents. We refer to this as the \textbf{\game} because it generalizes mechanism design in a number of ways; e.g., it involves voting on goals, and it involves the design of an economic policy for an economic environment in which agents  take actions and report types.
\def\outer{Voting Mechanism}
\def\inner{Stackelberg Game}

\subsection*{Further analysis and breakdown of Definition \ref{def:main-framework}.}

$\Theta$ is composed of $(n+1)$ vectors: $\Theta_{1}$ is the type space of the principal and $\Theta_{-1}$ to be the type space of the participants.  $\Theta$ can be added to the state space of the POMG, which allows dynamic types that change over time in response to the state of the game. We do not allow the Principal to directly manipulate or observe the state space. Thus, we can embed the type space within the state space to hide it from the Principal. Even with elements of the POMG that the Principal does have control over, such as the state transition function $T^{\phi}$, one can enforce hard constraints on how much power the Principal has to change the function explicitly through the divergence $D$ or implicitly through the implementation map $P$.

Both the infinite-horizon and finite-horizon version of the {\game} can be considered. In contrast to standard Reinforcement Learning (RL), we do not need a discount factor for the infinite-horizon version, as we consider the Principal as maximizing  the objective at the current voting round (and the objective may change at each round).
Exploring the tradeoffs between this local objective and more complex, sequential objectives is left as an important direction for future work. In the finite horizon case, we add an additional time horizon $\mathcal{T}$ to our game.

We now proceed to a detailed breakdown of our game. The \game~can be  divided into a \newterm{\outer} and \newterm{\inner}, which is played with the Principal's objective determined by the {\outer}.

\begin{definition}
    The \newterm{\outer} is defined as $\mathcal{V}=(\objectives, f, \Theta)$.
\end{definition}

Here, we use the standard axiomatic model \citep{arrow_social_2012}, where $\objectives$ is the set of alternatives, $f$ is the social choice function, and $\Theta$ is the type space (set of all preference profiles). A specific agent $i$'s type $\theta_i$ is a latent vector that represents  agent $i$'s values. This type contains all information necessary for recovering a partial ordering over alternatives. The goal of the {\outer} is  to define an objective for the Principal.  For this, we define the {\outer}, $f$, and ask the players for a preference report $\theta_{-1} \in \Theta_{-1}$, perhaps untruthful. We leave to future work whether some notion of approximate incentive compatibility can be achieved by the principal. The {\outer} then computes the objective $\objective = f(\theta_1, \theta_{-1})$ as a result of the vote, where $\theta_1 \in \Theta_1$ is the preferences of the Principal. By including the preferences of the principal, this objective function  allows expressing   a form of ``moral objectivity," or other biases.
It also allows to express mechanisms such as auctions, where the objective of the principal may be entirely selfish such as revenue,
and not depend at all on the participant's types.

\phantomsection\label{sec:obj-examples}
\paragraph{Social Welfare Examples.}
Examples of social welfare functions that could be included in the voting set are the Utilitarian objective:
\begin{equation}
    \objective(\phi, \pi) = \sum_{i} u_i(\phi, \pi).
\end{equation}
where $u_i(\phi, \pi)$ is as defined in \autoref{eqn:policy-objective}, $\pi$ is the tuple of all agents $\pi=(\pi_{i})_{i\in [n]}$, and $\pi_i$ maps $\Omega_i^{\phi} \to A_i^{\phi}$. Other possible choices include the Nash Welfare objective:
\begin{equation}
    \objective = \left(\prod_{i} u_i(\phi, \pi)\right)^{1/n}.
\end{equation}
and the Egalitarian objective:
\begin{equation}
    \objective = \min_i u_i(\phi, \pi).
\end{equation}
Custom welfare functions can also be considered.



\begin{definition}
    The \newterm{\inner}  is defined as    $I = (\Phi, P, D, \delta, \phi_0)$ and is a Stackelberg-Markov Game.
\end{definition}

The {\inner} game is played  after the {\outer}, and can be thought of as a single timestep of the full game. The Principal (leader) will choose action $\phi\in \Phi$ which induces a parameterized \newterm{\induced} $\pomg^{\phi}=(\states, \actions^{\phi}, \transfunc^{\phi}, \rewardfunc^{\phi}, \obs^{\phi}, \obfunc^{\phi}, \discount^{\phi}, \mu_0^{\phi})$ through the policy implementation map $P: \phi \mapsto \pomg^{\phi}$. If agent preferences change over time, this can be modeled by adding agent types into the state space of the POMG. The transition function $T$ would then be able to express changes in preferences over time.

The objective of the leader in the \inner~game is to design a POMG, given the objective $\objective$ as decided  in the {\outer}:~
\begin{equation}
    \begin{aligned}
        \quad               & \max_\phi \objective(\phi, \pi) \\
        \textrm{s.t.} \quad & D(\phi_0, \phi) \leq \delta     \\
        \textrm{and}\quad   & \mu_0^{P(\phi)} = \Delta(s_T).
    \end{aligned}
\end{equation} 

Again, $\pi$  is the tuple of all agents $\pi=(\pi_{i})_{i\in [n]}$, and $\pi_i$ individual agents that map $\Omega_i^{\phi} \to A_i^{\phi}$. Our notation $\mu_0^{P(\phi)}$ denotes the $\mu_0$ of the tuple $P(\phi)$, and $\Delta(s_t)$ refers to a Delta Dirac distribution centered on $s_T$. The second constraint $\mu_0^{P(\phi)} = \Delta(s_T)$ forces  $\phi$ to choose a POMG that has the same initial state as the terminal state of the last round so that continuity is kept between rounds.


This constrained optimization can  be transformed into an unconstrained problem by using an additional reparameterization $\mathcal{R}: \xi \mapsto \hat{\Phi}$, where $\xi \in \Xi := \mathbb{R}^L$ and $\hat{\Phi} := \{\phi \hspace{.25em} | \hspace{.25em} D(\phi_0, \phi) \leq \delta\}$.  The optimization can then proceed in $\mathbb{R}^L$ with no constraints. In this case, the Stackelberg game would reduce to $I = (\Xi, P')$, where $P' = P \circ \mathcal{R}$.
Finally, the \newterm{\induced} is defined as the POMG produced as the output of the principal.
\begin{definition}
    The \newterm{\induced} is  defined as $\pomg^{\phi}=(\states, \actions^{\phi}, \transfunc^{\phi}, \rewardfunc^{\phi}, \obs^{\phi}, \obfunc^{\phi}, \discount^{\phi}, \mu_0^{\phi})$ and is a Partially Observable Markov Game.
\end{definition}

Agents within the POMG interact with one another and attempt to maximize their utility according to their true preferences. The $n$  participants (followers)  play strategically in the parameterized POMG $\pomg^{\phi}$.
At each step $t$ of the game, each follower $i$ chooses an action $\actiont[i]$ from their action space $\actions[i]$, the game state evolves according to the joint action $\actiont=(\actiont[1], \cdots, \actiont[n])$ and the transition function $T$, and
agents receive observations and reward according to $\obfunc$ and $\rewardfunc$. An agent’s behavior
is characterized by its policy $\pi_i: \Omega_i^{\phi} \to A_i^{\phi}$, which maps observations to actions. Each follower   seeks to maximize its own (discounted) total return $\sum_{t} (\gamma^{\phi})^t r_i^{\phi}(s_t, \actiont[i], \actiont[-i])$.





\section{Example: Apple Picking Game}\label{sec:example}
\newcommand\figlength{.45}
\begin{figure}[ht!]
    \begin{center}
        \includegraphics[width=\figlength\textwidth]{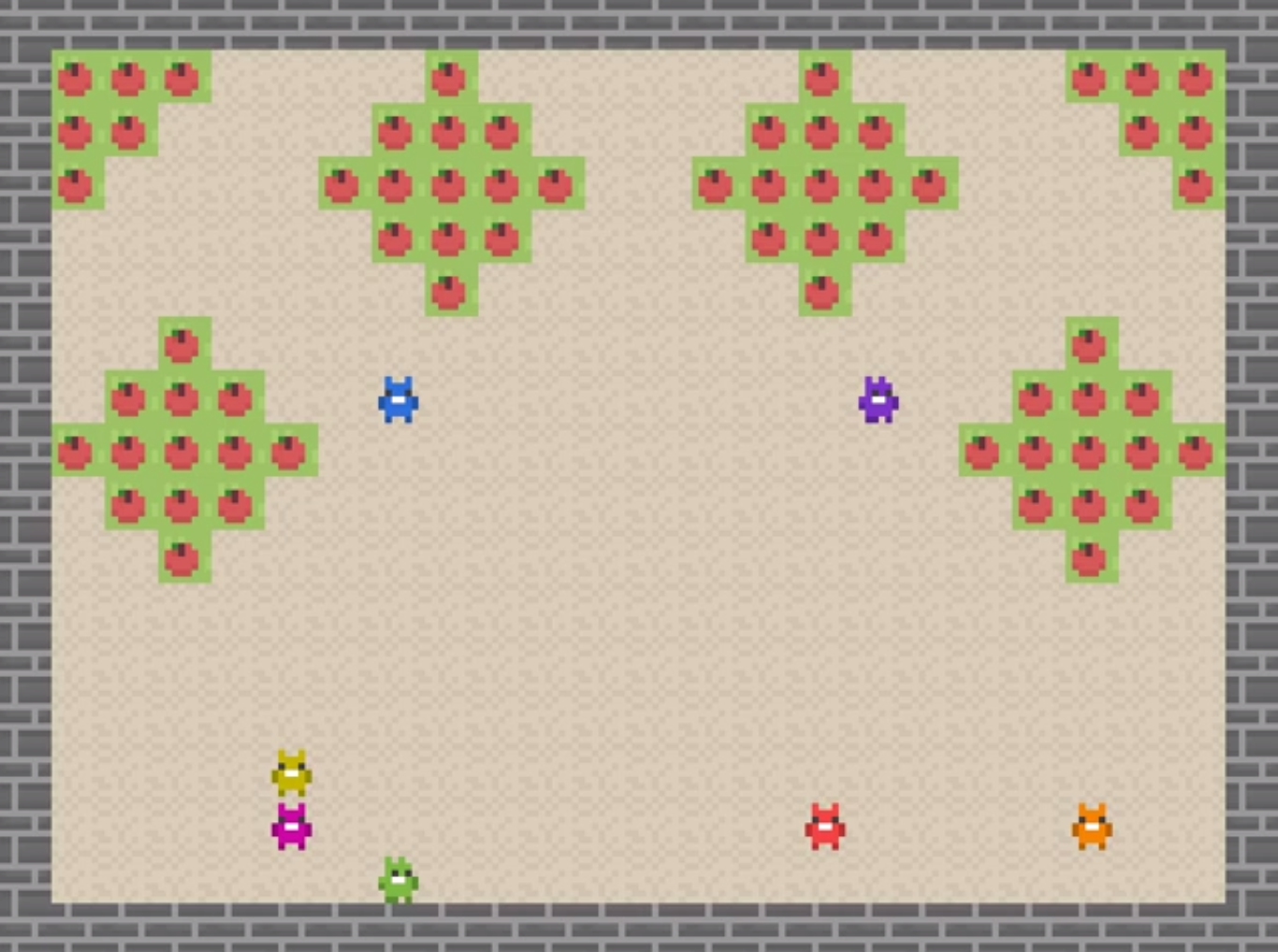}
    \end{center}
    \caption{An example of {\framework} as an apple picking game, built with Melting Pot 2.0 \citep{agapiou2022melting}. Player agents observe a restricted view of their environment, and receive a mixed reward depending on the apples they collect and the apples their observable neighbors collect. The principal observes both an unrestricted view of the environment and the running totals of all the players’ cumulative rewards, where it collects tax and redistributes wealth on the cumulated rewards at the end of every tax period (50 timesteps), similar to \citet{zheng2020ai}. Rewards for the Principal are determined by the change in value of the objective it is currently assigned by agent votes. For training details and hyperparameters, please refer to \autoref{apenndix:hp}.}\label{fig:social_dilemma}
    \vspace{-15pt}
\end{figure}

In order to give a motivating example for how preference elicitation for the principal in Social Environment Design can be used to align policy-maker incentives, we have created a Sequential Social Dilemma Game inspired by the \textit{Harvest Game}~\citep{perolat2017multiagent}. The aim in the Harvest game is to collect apples, with each apple yielding a reward. If all apples in an area are harvested, they never grow back. The dilemma arises when individual self-interest drives rapid harvesting, which could permanently deplete resources. Thus, agents must sacrifice personal benefit and cooperate for the collective well-being.

One potential solution to this dilemma is through the use of a central government that taxes and redistributes apples. Thus, we have created a new game in which a principal designs tax rates on apple collection, and players vote on Utilitarian (productivity) vs. Egalitarian (equality) objectives for the principal, similar to \citet{koster2022humancentered}. As players interact within this evolving environment, the principal faces the challenge of crafting policies that balance immediate economic incentives with sustainability goals. In order to achieve this, the principal must foster cooperation among players, guiding them towards the objective they have chosen. We release our code in the supplementary material for reproducibility.

To build intuition for how the Apple Picking Game maps onto the theoretical framework described in \autoref{sec:framework}, we give a more formal definition of the game here.  Recall the definition of a {\game} $\mathcal{S}=(\Phi, P, \phi_0, D, \delta, \Theta, \objectives, f)$. The action space for the environment designer $\Phi$ is defined as tax weights $0 \leq \Phi_i \leq 1$, determining the percentage of income to be taxed for each bracket. For simplicity we evenly redistribute the taxes, and set the number of tax brackets to three.

In this environment, the Principal can only change the reward function of the induced POMG, so $\pomg^{\phi}=(\states, \actions, \transfunc, \rewardfunc^{\phi}, \obs, \obfunc, \discount, \mu_0)$. Let there be $n$ agents, with the type of each agent defined by $\sigma$, or the selfishness of the agent. Finally, let $a_i$ be the number of apples collected for a given agent $i$. We can now define the policy implementation map $P$, which in this case reduces to the parameterization of the reward function. Here $a$ is a vector of all the apples collected with length $n$:

\begin{equation}
    r_i(a, \phi) = \sigma_i r_{\text{tax},i}(a) + (1 - \sigma_i)\left( \sum_{j \in n} r_{\text{tax}, j}(a, \phi) \right)
\end{equation}

The reward is an average between the apples an agent collects and the apples all other agents collect within its field of view weighted by the selfishness of the agent. The taxed reward is the amount of apples after tax an agent collects plus an equal share of the redistributed total tax:

\newcommand{\brck}[1]{\left(#1\right)}
\def\income{a}
\def\bracketcutoff{\tau}
\def\tax{T}
\def\taxrate{\phi}
\def\mtaxrate{\phi}
\def\taxperiodlen{M}
\def\incometax{ T }  %
\begin{align}
    r_{\text{tax}, i}(a, \phi)      & = (a_i - \tax(a_i, \phi)) + \frac{1}{n} \sum_{j \in n} \tax(a_j, \phi), \nonumber                                                          \\
    \text{where tax } \tax(a, \phi) & = \sum_{b=0}^{B-1} \mtaxrate_b \cdot ( \brck{ \bracketcutoff_{b+1} - \bracketcutoff_b } \bm{1}[ \income > \bracketcutoff_{b+1} ] \nonumber \\
                                    & + \brck{ \income - \bracketcutoff_b } \bm{1}[ \bracketcutoff_b < \income \leq \bracketcutoff_{b+1} ] ).  \nonumber
\end{align}

Here, $[\tau_b, \tau_{b+1}]$ refer to the tax brackets. Importantly, the principal can only incentivize agents through the taxed reward and cannot directly observe the true reward. We sample selfishness uniformly over [0, 1],  and keep them fixed during training. Allowing them to change over time either randomly or in some fashion dependent on the performance of the Principal is left for future work. The objective space of the Principal is defined as $\objectives = \{\eta \sum_{i,t} a_{i,t} + (1 - \eta) \min_{i} \sum_{t} a_(i,t) \hspace{.25em} | \hspace{.25em}  0 \leq \eta \leq 1\}$, an interpolation between the Utilitarian and Egalitarian objective. A simple social choice function $f(\sigma) = \eta$ can be defined as the average of agent selfishness: $f(\sigma) = \frac{1}{n} \sum_i \sigma_i$. In this setting, we leave the Principal optimization unconstrained and thus do not need to define $D$ or $\delta$. $\phi_0$ is initialized to $0$, or no tax.

We run several tax periods per voting round, and at the end of each voting round the principal decides on a new tax rate, for each bracket ---as well as calculating, applying and redistributing tax to the players for that entire period, delivered in the players’ final reward.  We release the codebase, which is designed for fast experimentation and further research, in the supplementary material, and include environment hyperparameters and training details in \autoref{apenndix:hp}. We do not include experimental results here, as the primary purpose of this paper is to propose a future research agenda and illustrate open problems.


\section{Challenges and Open Problems}

Based on the AI-led economic policy-making framework presented, the following key open problems of our framework are proposed for further exploration: \textbf{Preference aggregation and democratic representation} in voting mechanisms is a complex challenge that requires advanced algorithms to reflect collective preferences while respecting minority views, as well as ensuring that the simulated population is representative and their preferences correctly modeled. \textbf{Modeling human behavior} within the simulator is another key challenge, and points towards possibly  incorporating bounded-rationality into MARL \citep{wen2019modelling} or role-based modeling \citep{wang2020roma,wang2021rode}.~ To ensure responsible \textbf{AI governance and accountability}, responsible oversight mechanisms must be established. Furthermore, exploring socioeconomic interactions within these systems is important, especially in understanding and deriving the conditions for \textbf{convergence to and definition of the Principal's objective}. As our framework is positioned within a continual learning setting, it is important to redefine what an optimal Principal looks like in this context. Finally, \textbf{scaling laws} of the framework should be analyzed in order to fully model real-world complexities. Can the framework handle simulating economies with thousands or millions of agents? What is the role of scale? When is simulation useful, and when does it fail?

\textbf{Preference Aggregation and Democratic Representation.}

\textit{Aggregation algorithms within the {\outer}:} The development of sophisticated algorithms that can effectively aggregate disparate and potentially conflicting preferences of diverse agent populations is a significant challenge. These algorithms must ensure that the outcomes represent collective preferences without overwhelming the minority views.

\textit{Incorporating diverse decision-making models:} The framework must be flexible enough to respect various cultural, ethical, and socioeconomic decision-making paradigms that different groups of agents might exhibit. Such agents should imitate humans well, which we expand on further next.

\textbf{Modeling Human Behavior.}

\textit{Representative Agents:} Modeling agent behavior to accurately represent the diverse economic behaviors of real-world individuals is a complex and significant challenge. The agents should capture a variety of human traits, which include various decision-making styles, risk tolerance levels, and reactions to incentives and regulations.

\textit{Bounded Rationality:} Human decision-making in economic settings often demonstrates bounded rationality, where decisions are made based on satisficing rather than optimizing behavior. Further research is required to develop AI agents that can capture such nuances in human decision-making.

\textit{Agents' Perception of the System:} It may also be important for the follower AI agents in the framework to model how humans perceive the system overall, encompassing their beliefs about the function and credibility of the principal, the perceived power dynamics, and their understanding of collective objectives.

\textit{Cognitive and Behavioral Biases:} Human economic behavior is influenced by a variety of cognitive biases. For instance, time-inconsistent preferences can lead to procrastination and problems with self-control, and loss aversion can skew risk preferences. AI agents within this framework need to capture these biases for accurate representation of human economic behavior.

\textit{Interaction and Network Effects:} Humans do not make economic decisions in isolation; their decisions are  influenced by their interactions with others. This opens another avenue of research in modeling these network effects accurately within the agent behavior models. Higher order effects are oftentimes essential to understanding the behavior real-world systems.

\textit{Role-Based Modeling:} Finally, research on modeling the behavior and decisions of real-world policymakers and the influences shaping their choices based on their role is essential \cite{wang2020roma}. This is crucial for the principled design of the principal agent that designs economic policies.

\textbf{AI Governance and Accountability.}

\textit{Transparent decision-making processes:} AI systems involved in policy-making benefit from transparent decision-making processes. We view the creation of interpretable AI models that can provide explanations for suggested policies is essential for trust and accountability.

\textit{Legal and ethical frameworks for AI decisions:} There is a need to establish legal and ethical frameworks that delineate the responsibilities and liabilities associated with AI-driven decision-making. These frameworks should set guidelines for what constitutes fair and lawful AI behavior in an economic context.

\textit{Oversight and human-AI collaboration:} Establishing effective oversight mechanisms that involve both AI and human collaboration is important. The role of human experts in supervising and guiding AI decisions, and their ability to intervene when AI-driven policies deviate from desired outcomes, is still to be determined.

\textbf{Convergence to Desired Outcomes.}

\textit{Existence and characterization of forms of convergence or equilbria:} Can we characterize the conditions under which an equilibrium will exist in such complex socioeconomic interactions? The uniqueness or multiplicity of equilibria and the conditions  under which they are attained are also interesting to study. Also, conventional game-theoretic equilibria may not be the right object of study, as empirically these economic systems may never converge to a single, stable behavior.

\textit{Algorithmic stability and multi-agent coordination:} Identifying reinforcement learning algorithms that can demonstrably converge is another open problem. In addition, coordination among several agents, with varying objectives and possibly divergent strategies, also remains a large open problem.

\textit{Influence of dynamic changes on convergence points:} The complex dynamics of economic systems call for a deep understanding of the sensitivity of equilibria to shocks and changes in the environment and agent behavior from variables that may have been unforeseen by the principal. Ensuring the robustness and stability of the principal to be able to recover from such shocks is also of importance.

\textbf{Scaling Laws and Computational Efficiency.}

\textit{Scaling up the model to larger systems:} The proposed framework needs to be scaled to simulate economies of increasingly complexities. This comprises accommodating an increasing number of agents and more intricate interactions among them. Scaling laws of the model parameters and the computational resources required need to be examined.

\textit{Efficient learning and decision-making algorithms:} Efficient algorithms for learning agent behavior and optimizing the policy design are crucial for the practicality of the framework. Particularly, the principal must be sample efficient, as every step it takes induces an entire MARL optimization.

\textit{Massive parallelization:} To tackle real-world complex systems, embracing the advantage of high-performance computing is necessary. This includes implementing the framework with massively parallel computations for both the learning and the decision-making processes. Techniques for splitting these processes into smaller tasks that can be processed simultaneously, as well as the efficient management of these tasks, represent challenging aspects to be addressed.

\textit{Model compression techniques:} Consideration and application of model reduction techniques can be crucial for simulation feasibility. It is important to identify the dominant features and behaviors that drive the system outcome, and focus computational resources on those, while simplifying or neglecting less influential details.

\textit{Robustness and performance evaluation:} The robustness of the Principal model and follower models with respect to the increase in scale, and the degradation in performance under computational constraints should be examined in future work. This includes an analysis of the trade-off between model performance and computational efficiency.

\section{Related Work}\label{sec:related}

The concept of environment design was first proposed by~\citet{zhang2009general} and focused on the single agent setting. In contrast, our framework resides between various strands of research, including but not limited to economic policy design, Stackelberg game learning, multi-agent reinforcement learning, mechanism design, and computational social choice. In this section, we delve into a comprehensive exploration of its connections with prior research.

\subsection{Economic Policy Design and Simulation}

Several approaches to automated economic policy design have been proposed in the past \citep{pmlr-v162-liu22l,curry2023learning,yang2021adaptive}, and how usage of AI may span both participation in and design of economic systems \citep{parkes2015reasoning}. Here we cite several that are most related to our proposed framework and research agenda. Perhaps most related to our approach is  Human Centered Mechanism Design~\citep{koster2022humancentered, balaguer2022hcmd}. They propose learning mechanisms from behavioral models trained on human data, with the mechanism objective attempting to satisfy a majoritarian vote of the human participants. However, their work differs from ours in several key ways; firstly, they do not consider a fully general economic environment and limit their scope only to a generalization of the linear public goods setting. In other words, our framework encompasses Environment Design whilst theirs encompasses only Mechanism Design. Secondly, the voting that is defined within their framework is taken over actual mechanisms proposed by the designer and is by majority, whereas our voting is taken explicitly over Principal objectives and does not specify a majority vote, which allows potentially addressing issues such tyranny of the majority. A more general game environment is illustrated in the AI Economist~\citep{zheng2020ai} and its application to taxation policy, although they adopt a Principal with a fixed goal that is not subject to voting by participants.

\subsection{Stackelberg Game}

From the perspective of the Principal, it plays a Stackelberg game with agents of different types. Stackelberg games model many real-world problems that exhibit a hierarchical order of play by different players, including taxation~\cite{zheng2020ai}, security games~\cite{jiang2013defender, gan2020mechanism}, and commercial decision-making~\cite{naghizadeh2014voluntary, zhang2016multi, aussel2020trilevel}. In the simplest case, a Stackelberg game contains one leader and one follower. For these games with discrete action spaces,~\citet{conitzer2006computing} show that linear programming approaches can obtain Stackelberg equilibria in polynomial time in terms of the pure strategy space of the leader and follower. To find Stackelberg equilibria in continuous action spaces, \citet{jin2020local, fiez2020implicit} propose the notion of local Stackelberg equilibria and characterize them using first- and second-order conditions. Moreover,~\citet{jin2020local} show that common gradient descent-ascent approaches can converge to local Stackelberg equilibria (except for some degenerate points) if the learning rate of the leader is much smaller than that of the follower.~\citet{fiez2020implicit} give update rules with convergence guarantees. Different from these works, in this paper, we consider Stackelberg games with multiple followers.

More sophisticated than its single-follower counterpart, unless the followers are independent \citep{calvete2007linear}, computing Stackelberg equilibria with multiple followers becomes NP-hard even when assuming equilibria with a special structure for the followers \citep{basilico2017methods}. \citet{wang2021coordinating} propose to deal with an arbitrary equilibrium which can be reached by the follower via differentiating though it. \citet{gerstgrasser2023oracles} proposes a meta-learning framework among different policies of followers to enable fast adaption of the principal, which builds upon prior work done by \citet{brero2022learning} who first introduced the Stackelberg-POMDP framework. \citet{hossain2024multi} study the multi-sender persuasion game as a special case of Stackelberg game with multiple principals.

Multi-agent reinforcement learning holds the promise to extend Stackelberg learning to more general and realistic problems.~\citet{tharakunnel2007leader} propose Leader-Follower Semi-Markov Decision Process to model the sequential Stackelberg learning problem.~\citet{cheng2017multi} propose Stackelberg Q-learning but without any convergence guarantee.~\citet{shu2018m, shi2019learning} study leader-follower problems from an empirical perspective, where the leader learns deep models to predict the followers' behavior.

\subsection{Multi-Agent Reinforcement Learning}

Another component of the proposed framework is the followers' behavior learning. Deep multi-agent reinforcement learning~\cite{yu2022surprising,wen2022multi,kuba2021trust,christianos2020shared,peng2021facmac,jiang2019graph,rashid2018qmix,dong2022low,dong2023symmetry,wang2019influence} algorithms have seen considerable advancements in recent years. Notable contributions such as COMA \citep{foerster2018counterfactual}, MADDPG \citep{lowe2017multi}, PR2 \citep{wen2019probabilistic}, and DOP \citep{wang2021off} address policy-based MARL challenges. These approaches leverage a (decomposed) centralized critic for computing gradients to decentralized actors. Conversely, value-based algorithms decompose the joint Q-function into individual Q-functions, facilitating efficient optimization and decentralized implementation. Techniques such as VDN \citep{sunehag2018value}, QMIX \citep{rashid2018qmix}, QTRAN \citep{son2019qtran}, and Weighted QMIX \citep{rashid2020weighted} incrementally enhance the mixing network's representational capacity. Additional investigations explore MARL through coordination graphs \citep{guestrin2002coordinated, guestrin2002multiagent, bohmer2020deep,kang2022non,wang2021context,yang2022self}, communicative strategies \citep{singh2019learning, mao2020neighborhood,wang2019learning,kang2020incorporating}, diversity~\citep{li2021celebrating}, and also expressive neural network architectures like Transformer~\citep{wen2022multi}, offering insights for participant learning without directly addressing human behavior modeling.

For modeling behavior, role-based learning frameworks~\cite{wang2020roma, wang2021rode} are the most related to our work. They learn the roles of different agents autonomously and enhance learning efficiency by decomposing the task and learning sub-task-specific policies. However, these works are mainly studied in the setting of the Decentralized Partially Observable Markov Decision Process (Dec-POMDP), and are thus differ from the present work in two ways: (1) The reward is shared among agents; and (2) The dynamics, including reward and transition dynamics, is fixed. There likely would exist significant challenges in generalizing these to the kinds of non-shared reward settings that are essential for many economic applications.

\subsection{Computational Social Choice}


Computational social choice is an interdisciplinary field combining computer science and social choice theory, focusing on the application of computational techniques to social choice mechanisms (such as voting rules or fair allocation procedures) and the theoretical analysis of these mechanisms with computational tools \citep{brandt2015handbook}. A fundamental component of the field is the study of manipulative behavior in elections and other collective decision-making processes, as well as the design of systems resistant to manipulation \citep{elkind2015rationalizations, procaccia2010approximation}. This area of study will likely inform the development of the  Voting Mechanism. Additionally, computational social choice attempts to optimize the fair distribution of resources, often involving complex allocation problems \citep{thomson2015theory,procaccia2015cake}.

The present work deviates from typical computational social choice models in that it attempts to simulate the actual policy outcomes of the vote within a simulated economy, which is a departure from the analysis typically found in computational social choice. While this represents a step towards greater real-world practicality, it also dramatically increases both the difficulty and complexity of the problem setting.

\subsection{Automated Mechanism Design}

Automated Mechanism Design (AMD)~\citep{sandholm_automated_2003}
makes use of search algorithms for the design of
specific rule sets (mechanisms) for games that lead to desirable outcomes even when participants act in self-interest. Our work shares some of the motivations of automated mechanism design (AMD) in that it hypothesized that automated approaches would someday outperform traditional manual designs, be applicable to a broader range of problems, and circumvent economic impossibility results, by transferring the burden of design from humans to machines.

The work of AMD has also been advanced through deep learning in the framework known as \textit{differentiable economics}. \citet{duetting2023optimal} use deep neural networks to learn the allocation and payment rules of auctions. Since then, a line of follow-up work has been introduced, extending the framework to make the architecture more powerful and general \citep{shen_automated_2021,ivanov2022optimal,duan2023scalable,curry2022learning,wang2023deep,wang24x}. Deep learning methods have also been explored in equilibrium calculation~\citep{kohring2023enabling,bichler2023learning,bichler2021learning}. While these techniques are applied to settings  less general than ours, some of the architectural details may be useful in building a Principal.

\section{Conclusion}\label{sec:conclusion}

In this paper, we have presented a framework for policy design and simulation that merges economic policy design with AI to potentially help better inform economic policy-making. It is designed to tackle issues such as preference aggregation and counterfactual testing in complex economic systems. Significant challenges, including democratic representation and accountability in AI-driven systems, are highlighted.  We hope to engage interdisciplinary expertise and foster collaborative innovation, and aspire to help create AI systems that not only enhance economic resilience and governance effectiveness but also uphold democratic ideals and ethical standards.


\section*{Acknowledgements}
We would like to express enormous gratitude to David Bell for providing many hours of sustained feedback and support throughout the process of writing, and his belief in the framework's potential. We would also like to thank Austin Milt, and Benjamin Smith for their dedication of several hours of analyzing, critiquing, and improving the framework. Finally, we would like to thank Rosie Zhao, Gianluca Brero, Gabriel Totev, Ariel Procaccia, Dima Ivanov, and Matteo Bettini for reading the draft and providing feedback on improving the paper.

\clearpage
\section*{Impact Statement}
This paper sets out an agenda in {\framework}, suggesting that AI holds promise in improving policy design by proposing a general framework that can simulate general, socioeconomic phenomena and scale to large settings. There are several relevant  considerations that are important to take-up in advancing this framework towards adoption by policy-makers. For example,  its effectiveness depends on capturing all pertinent stakeholders within a given scenario. Related, is to ensure
that  agent modeling is consistent with the diverse motivations and incentives of people, firms, and other entities. Lastly, any real-world trial of this initiative should engage vigorously and faithfully with non-technical stakeholders.

\bibliography{main}
\bibliographystyle{icml2024}

\newpage
\appendix
\onecolumn

\section{Environment Hyperparameters and Training Details}\label{apenndix:hp}

Here we give a detailed breakdown of several key hyperparameters and Training Details within our environment in \autoref{sec:example}.

We use PPO \cite{schulman2017proximal} player agents with parameter sharing and GAE \cite{schulman2015high}, collecting samples at a horizon shorter than the episode length to perform multiple policy update iterations per episode. The principal has separate, discrete, action subspaces for each tax bracket, and is also trained by standard PPO at the same time-scale as the player agents. We follow a two-phase curriculum with tax annealing, as suggested in \citet{zheng2020ai}. This annealing can be formalized as a constraint in the policy implementation map by simply bounding the maximum tax percentage that can be set. It is worth noting, however, that training the principal in this way is susceptible to issues of non-stationarity, and we refer to \citet{yang2021adaptive} for a discussion on alternatives.

To give a further explanation regarding the Apple Respawn Probabilities, the probability of a respawn per timestep depends on how many neighbors are around it in a circular radius of 2. With four neighbors, the respawn probability is $0.025$. With 0, the probability becomes $0$.


\begin{table}[h]
    \vspace{10pt}
    \centering
    \begin{tabular}{lll}
        \toprule
         & Hyperparameter                               & Value                                                \\
        \midrule

         & Number of Agents                             & $7$                                                  \\
         & Initial Number of Apples                     & $64$                                                 \\
         & Apple Respawn Probabilities                  & [$0.025$, $0.005$, $0.0025$, $0.0$]                  \\
         & Base Reward                                  & $1$ on apple collection                              \\
         & Social Reward                                & $1$ on apple collection of observable agents         \\
         & Agent Type $(\sigma, \beta)$                 & Sampled from $\text{Uniform} \left[ 0, 1 \right]$    \\
         & Agent Observability (units are grid tiles)   & (Forward: $9$, Right: $5$, Backward: $1$, Left: $5$) \\
         & Principal Tax Brackets (units are in apples) & [($1$,$10$),($11$,$20$),($21$,$10000$)]              \\
         & Tax Period                                   & $50$                                                 \\
         & Episode Length                               & $1000$                                               \\
         & Sampling Horizon                             & $200$                                                \\
        \bottomrule
    \end{tabular}
    \caption{Hyperparameters for our methods in \autoref{sec:example}.}
    \label{tab:mujoco-hp}
\end{table}

\end{document}